# Towards Generalized Hydrological Forecasting using Transformer Models for 120-Hour Streamflow Prediction

Bekir Z. Demiray[a]* and Ibrahim Demir[a,b,c]

[a] IIHR—Hydroscience and Engineering, University of Iowa, Iowa City, Iowa, USA
[b] Civil and Environmental Engineering, University of Iowa, Iowa City, Iowa, USA
[c] Electrical and Computer Engineering, University of Iowa, Iowa City, Iowa, USA

* Corresponding Author, Email: bekirzahit-demiray@uiowa.edu

**Abstract**
This study explores the efficacy of a Transformer model for 120-hour streamflow prediction across 125 diverse locations in Iowa, US. Utilizing data from the preceding 72 hours, including precipitation, evapotranspiration, and discharge values, we developed a generalized model to predict future streamflow. Our approach contrasts with traditional methods that typically rely on location-specific models. We benchmarked the Transformer model's performance against three deep learning models (LSTM, GRU, and Seq2Seq) and the Persistence approach, employing Nash-Sutcliffe Efficiency (NSE), Kling-Gupta Efficiency (KGE), Pearson's r, and Normalized Root Mean Square Error (NRMSE) as metrics. The study reveals the Transformer model's superior performance, maintaining higher median NSE and KGE scores and exhibiting the lowest NRMSE values. This indicates its capability to accurately simulate and predict streamflow, adapting effectively to varying hydrological conditions and geographical variances. Our findings underscore the Transformer model's potential as an advanced tool in hydrological modeling, offering significant improvements over traditional and contemporary approaches.

## 1. Introduction

Recent years have seen a significant escalation in both the frequency and severity of natural disasters across the globe, underscoring the need for more reliable forecasting tools that extend beyond immediate alerts. According to an analysis by the World Meteorological Organization (2021), it is estimated that weather, climate, or water-related disasters have on a daily average over the last fifty years, resulted in financial losses of $202 million and led to 115 fatalities. Additionally, a 2022 report by Munich Re highlights that natural catastrophe, which include a range of disasters from hurricanes to floods, have cumulatively caused over $280 billion in damages worldwide. Specifically, in the United States, these disasters have led to an estimated $145 billion in damages, not to mention the immense loss of lives and significant property destruction. Research continues to point towards climate change as a driving force behind the increasing intensity and frequency of these extreme natural events, predicting a rise in both the scale of destruction and the number of affected individuals (WMO, 2021; UNESCO, 2023; Banholzer et al., 2014; IPCC, 2022).

Flooding ranks as the most frequent natural disaster, causing extensive financial damages and claiming numerous lives (WHO, 2021). Notably, in 2020, floods constituted over 60% of all natural disasters reported, contributing to 41% of the total deaths from such events (NDRCC, 2021). Research indicates that climate change is exacerbating the frequency and severity of flooding in various regions (Davenport et al., 2021; NOAA, 2022; Tabari, 2020; IPCC, 2022). This trend can be attributed to a variety of factors, including rising sea levels (Strauss et al., 2016), increased extreme rainfall events (Diffenbaugh et al., 2017), and more intense rainfall during hurricanes (Trenberth et al., 2018). Therefore, accurate streamflow forecasting is critical in effectively minimizing the impact of floods, particularly in terms of property damage and loss of life (Alabbad and Demir, 2022).

Additionally, streamflow forecasting serves as a cornerstone for informed decision-making in various areas of hydrology and water resource management. Applications ranging from watershed conservation (Demir and Beck, 2009) to precision agriculture (Yildirim et al., 2022) and the development of effective flood mitigation strategies (Li and Demir, 2022; Ahmed et al., 2021; Yaseen et al., 2018) heavily rely on this data. However, the complex and dynamic nature of hydrological systems, often characterized by nonlinear patterns and variability, makes achieving accurate streamflow predictions a persistent challenge (Honorato et al., 2018; Yaseen et al., 2017; Sit et al., 2024).

Numerous physical and data-centric methodologies focused on streamflow forecasting have emerged over time, each distinct in its approach, ranging from the use of varied data types to focusing on specific regions or levels of generalization (Salas et al., 2000; Yaseen et al., 2015). Approaches driven by physical principles (Beven and Kirkby, 1979; Ren-Jun, 1992; Arnold, 1994; Lee and Georgakakos, 1996; Devia et al., 2015) are capable of modeling complex physical interactions, including atmospheric dynamics and the evolution of global weather patterns (Yaseen et al., 2019; Sharma and Machiwal, 2021). While these models are invaluable, their implementation presents constraints, such as the need for detailed hydrological and geomorphological data, which leads to increased operational expenses. Additionally, their predictive accuracy tends to diminish for long-term forecasting.

Compounding these limitations, traditional physics-based models, due to their high computational demands and extensive parameter requirements, necessitate significant computational resources, resulting in notable costs (Mosavi et al., 2018; Sharma and Machiwal, 2021; Liu et al., 2022; Castangia et al., 2023). In response, recent studies have shifted focus to alternative methods, particularly highlighting the efficacy of machine learning and deep learning models as capable and often more accurate alternatives to physical models (Yaseen et al., 2015). Deep learning models, in particular, have demonstrated considerable improvements in both the accuracy and dependability of streamflow predictions, thus offering a potential transformative impact on the field of hydrological modeling (Demiray et al., 2023; Sit et al., 2023).

While traditional machine learning techniques, like Support Vector Machines (SVMs) and Linear Regression (LR), have found applications in streamflow forecasting and environmental analysis (Bayar et al., 2009; Li and Demir, 2024; Granata et al., 2016; Yan et al., 2018; Sharma and Machiwal, 2021), breakthroughs in artificial intelligence (AI) and enhanced GPU capabilities have catapulted the development of deep learning, ushering in new possibilities for this field (Sit et al., 2022). Among the diverse array of neural network architectures considered for this purpose (Sit et al., 2021; Chen et al., 2023), Recurrent Neural Networks (RNNs), particularly Long Short-Term Memory (LSTM) networks and Gated Recurrent Units (GRUs), stand out as the most extensively studied and applied models in this field.

Kratzert et al. (2018) has applied LSTM models to daily runoff prediction, integrating meteorological data and finding that LSTM models can surpass the efficacy of established physical models in specific areas. Similarly, Xiang et al. (2021) found that LSTM-seq2seq models outperform linear models like linear regression, lasso regression, and ridge regression in terms of predictive accuracy and other evaluation metrics. Furthermore, Guo et al. (2021) conducted a comparative study of LSTMs, GRUs, and SVMs across 25 different locations in China, concluding that LSTMs and GRUs show similar performance levels, with GRUs offering the advantage of quicker training times. To gain a broader perspective on the depth of research in deep learning applications for streamflow prediction, extensive studies such as those by Yaseen et al. (2015) and Ibrahim et al. (2022) provide valuable insights.

Originally pioneered for language translation by researchers at Google (Vaswani et al., 2017), the Transformer architecture has been explored for diverse tasks, including time series analysis (Zhou et al., 2021; Wu et al., 2021; Zhou et al., 2022; Lin et al., 2022). Transformer applications for streamflow forecasting, however, remain a relatively new and actively developing research area. For instance, in the Mekong River Basin, a study demonstrated that LSTM models perform better than Transformers, particularly in dry seasons (Nguyen et al., 2023). Another study utilized a transfer learning approach on Transformers for effective flood prediction in data-sparse basins of the Yellow River, showcasing its potential in areas with limited data (Xu et al., 2023).

Furthermore, Liu et al. (2022) developed a Transformer-based model for monthly streamflow prediction on the Yangtze River, demonstrating its ability to incorporate both historical water levels and the influence of ENSO patterns. Similarly, Castangia et al. (2023) applied a Transformer model for predicting daily water levels within a river network, with a focus on capturing upstream hydrological signals. They successfully evaluated this technique using data from the 2014 Southeast Europe flood event. Additionally, research comparing various deep-learning models in streamflow prediction, including Transformers, has shown that different preprocessing and data extension methods can significantly affect predictive outcomes (Demiray et al., 2024), emphasizing the importance of model selection and data handling in hydrological forecasting.

In this study, we expand the scope of streamflow forecasting by utilizing a Transformer model to predict water levels for the next 120 hours across 125 different locations within Iowa, US. Our methodology involves using data from the preceding 72 hours, encompassing precipitation, evapotranspiration, and discharge values, as well as the location-based features such as slope and soil types to forecast future streamflow. A unique aspect of this study is the development of a generalized model, trained on data from all 125 stations simultaneously, rather than creating individual models for each location. The performance of this unified Transformer-based model is then compared with three other deep learning models as well as the traditional persistence method. The experimental results demonstrate that our Transformer-based approach notably surpasses all other methods in forecasting accuracy, showcasing its effectiveness in handling a broad range of hydrological data across multiple locations.

This paper is organized as follows: The forthcoming section will introduce the dataset employed in this research along with a description of the study area. Section 3 is dedicated to detailing the methodologies implemented in this study. Subsequently, Section 4 will unveil the experimental results, accompanied by an in-depth analysis and discussion of these findings. Finally, Section 5 will encapsulate the principal conclusions drawn from this research and explore potential avenues for future work.

## 2. Case Study and Dataset

WaterBench-Iowa, developed by Demir et al. (2022), is the foundational dataset for this study, specifically designed as a benchmark dataset to streamline hydrological forecasting research. This dataset, in alignment with the FAIR (findability, accessibility, interoperability, and reuse) principles, offers a valuable resource for data-driven and machine learning applications in streamflow forecasting. It encompasses an expansive collection of hydrological and meteorological data from 125 distinct locations across Iowa, gathered from reputable sources including NASA, NOAA, USGS, and the Iowa Flood Center. The location of 125 sensors in Iowa is depicted in Figure 1. The dataset covers a comprehensive period from October 2011 to September 2018, providing a rich temporal dataset for our analysis.

The WaterBench-Iowa dataset integrates various key hydrological data, including streamflow measurements, precipitation records, watershed characteristics, slope, soil types, and evapotranspiration. This broad spectrum of data allows for a deepened understanding of hydrological processes across diverse environmental conditions. The dataset's structure is conducive to a variety of machine learning and deep learning methodologies, with a high resolution in both temporal and spatial dimensions, along with extensive metadata and relational information.

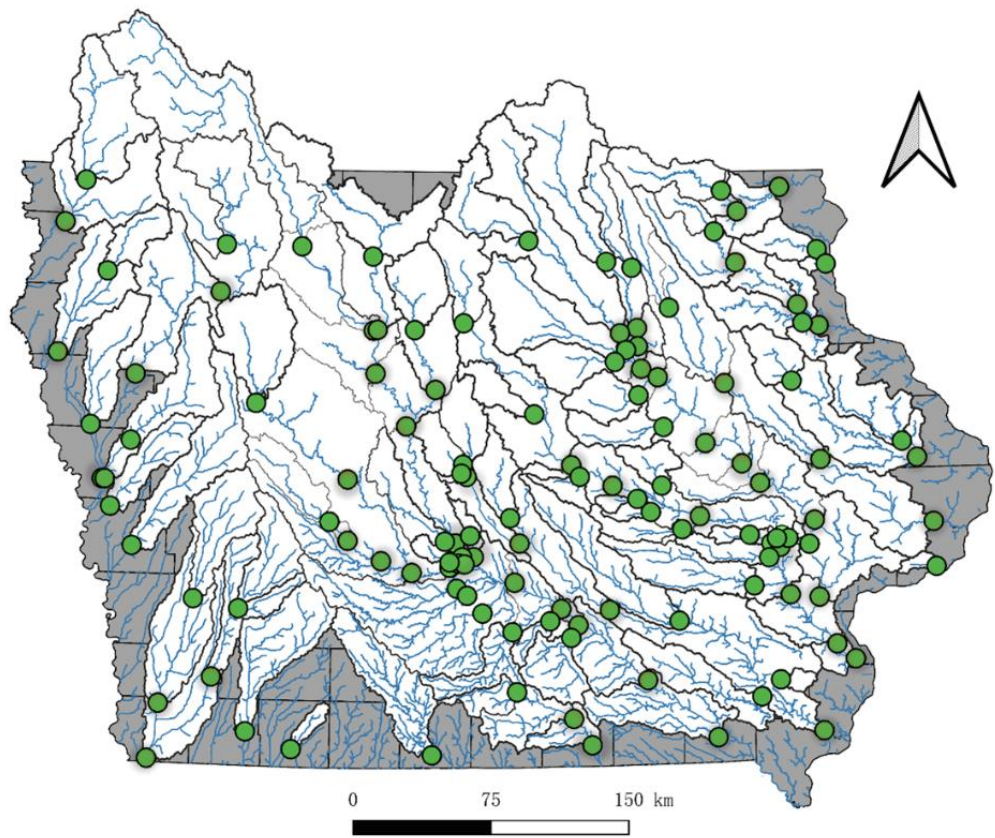

Figure 1: Location of used sensors in Iowa, US (adapted from Demir et al., 2022)

In this study, we employed the most recent water year in the dataset as the test set, with the remaining data used for training. Additionally, 15% of the training set was allocated as a validation set, ensuring robust model evaluation and fine-tuning. Preprocessing methods in this study replicate those established in the WaterBench-Iowa publication (Demir et al., 2022). This adherence ensures the validity of comparisons between our results and the dataset's existing benchmark models. A statistical summary of the dataset is provided in Tables 1 and 2. Our research approach involved the use of a generalized model, trained using data from all 125 stations. This approach marks a departure from traditional practices that often rely on location-specific models. The generalized model's ability to simultaneously process data from multiple locations not only demonstrates its versatility but also underscores its potential in offering comprehensive insights into streamflow forecasting.

Table 1: Statistical summary of watershed characteristics for 125 USGS gauges in the state of Iowa

|  | Area (km$^2$) | Concentration time (h) | Slope | Loam | Silt | Sandy clay loam | Silty clay loam |
|---|---|---|---|---|---|---|---|
| Min | 6 | 2 | 0.38% | 0% | 0% | 0% | 0% |
| Max | 36453 | 315 | 4.32% | 98% | 100% | 84% | 93% |
| Mean | 5405 | 77 | 1.97% | 33% | 31% | 18% | 18% |
| Median | 1918 | 53 | 1.80% | 33% | 21% | 4% | 7% |
| SD | 8320 | 68 | 0.80% | 28% | 30% | 24% | 23% |

Table 2: Summary statistics of precipitation and streamflow across 125 catchments from water year 2012–2018, including missing data analysis

|  | Annual total precipitation (mm) | Max. hourly precipitation (mm) | Annual mean streamflow (m3/s) | Missing rate of precipitation (raw data) | Missing rate of streamflow (raw data) |
|---|---|---|---|---|---|
| Min | 794 | 9.1 | 3 | 0.02% | 0.69% |
| Max | 1056 | 60.0 | 12963 | 0.04% | 33.14% |
| Mean | 952 | 24.8 | 1926 | 0.02% | 15.16% |
| Median | 961 | 22.2 | 608 | 0.02% | 16.14% |
| SD | 57 | 10.3 | 2864 | 0.01% | 6.4% |

## 3. Methods

In this research, we focused on assessing the performance of a Transformer-based model for predicting streamflow and conducted a comparative analysis with four other models: Persistence, GRU, LSTM, and Seq2Seq, as referenced in the WaterBench dataset. This section will elaborate on the methodologies applied in this study, detailing both the Transformer model and the comparative models.

### 3.1. Persistence Method

The Persistence approach (Eq. 1), often referred to as the nearest frame method, operates on the premise that future conditions will mirror the present. This method posits that the upcoming streamflow will closely resemble the most recent observations, essentially predicting that 'tomorrow will be the same as today'. Widely recognized as a baseline in hydrological research, including streamflow forecasting, the Persistence model holds a significant place, particularly in short-range forecasting scenarios. Numerous studies in hydrology (Krajewski et al., 2021) have noted that surpassing the accuracy of the Persistence model can be a challenge. This challenge is especially pronounced when predictions are made for periods shorter than 12 hours.

$$\hat{Y}_{i+n} = Y_i \qquad \text{Eq. 1}$$

$$\hat{Y}_{i+n} = Predicted\ streamflow\ values\ between\ time\ t\ and\ t + n$$
$$Y_i = Observed\ streamflow\ values\ at\ time\ t$$

### 3.2. Long Short-Term Memory

In time-series forecasting, while Recurrent Neural Networks (RNNs) have been pivotal in recognizing temporal patterns, they are often hindered by the vanishing gradient issue, limiting their capability in handling long-term dependencies. This challenge can significantly affect the precision of time-series forecasts, particularly in scenarios requiring memory of distant past events. To overcome these limitations, Long Short-Term Memory (LSTM) networks (Hochreiter and Schmidhuber, 1997) were developed. LSTMs are adept at maintaining short-term memory over extended periods and efficiently processing long-term data dependencies. This makes them particularly suitable for time-series forecasting and hydrological tasks like flood and rainfall prediction, as evidenced by various studies (Kratzert et al., 2018; Feng et al., 2020; Frame et al., 2022; Sit et al., 2022).

An LSTM node receives an input tensor $x_t$ and a hidden state tensor $h_{t-1}$ from the preceding node. Within the cell, multiple gates regulate information flow and memory retention. The input gate ($i_t$) controls the extent of new information to be stored, using weight matrices $W(i)$ and $U(i)$. It does this through a sigmoid function based on linear transformations of $x_t$ and $h_{t-1}$. Similarly, the forget gate ($f_t$) determines information to discard from the cell state, employing a sigmoid function with weight matrices $W(f)$ and $U(f)$. The output gate ($O_t$) dictates how much of the cell state to expose as the hidden state, utilizing a sigmoid function with weight matrices $W(O)$ and $U(O)$ also influenced by $x_t$ and $h_{t-1}$. The candidate cell state ($\tilde{c}$), signifying potential additions to the cell state, is calculated using the hyperbolic tangent function ($tanh$) and linear transformations of $x_t$ and $h_{t-1}$, with weight matrices $W(c)$ and $U(c)$.

The cell state $c_t$ is then updated, incorporating inputs from the forget gate, input gate, and candidate cell state, as illustrated in Equation 2:

$$c_t = f_t \odot c_{t-1} + i_t \odot \tilde{c}_t \qquad \text{Eq. 2}$$

Subsequently, the hidden state $h_t$ is generated by applying the output gate to the hyperbolic tangent of the updated cell state, as shown in Equation 3:

$$h_t = O_t \odot \tanh(c_t) \qquad \text{Eq. 3}$$

This updated hidden state $h_t$ and cell state $c_t$ are then relayed to the next LSTM node in the sequence and to subsequent layers within the neural network. In hydrological forecasting, LSTM networks have demonstrated superior performance compared to basic RNNs and other time-series forecasting models, garnering popularity in the field. By effectively resolving the vanishing gradient problem and capturing intricate long-term dependencies, LSTMs have established themselves as invaluable tools for precise and dependable predictions in various hydrological scenarios.

### 3.3. Gated Recurrent Units

Gated Recurrent Units (GRUs), introduced by Cho et al. (2014), represent a compelling alternative to Long Short-Term Memory (LSTM) networks for time-series forecasting. With their simplified internal structure, GRUs address efficiency concerns caused by LSTM complexity, making them advantageous for large-scale datasets common in hydrological forecasting.

GRUs achieve this through a dual-gate mechanism. The update gate ($z_t$) plays a crucial role in determining the extent to which new input data ($x_t$) and the previous hidden state ($h_{t-1}$) should contribute to the formation of the current hidden state ($h_t$). This gate enables the GRU to weigh the importance of recent versus historical information continually. In parallel, the reset gate ($r_t$) offers a powerful mechanism for discarding irrelevant past data, allowing the model to adapt to new trends and anomalies effectively. This selective forgetting is essential for maintaining model accuracy, especially in dynamic and complex hydrological environments where past trends may not always be indicative of future patterns.

The candidate hidden state ($\check{h}_t$), generated as a function of the reset gate, captures new information at the current timestep, considering only the most pertinent historical context. The final hidden state for the next step in the network is a blend of this candidate state and the past hidden state, modulated by the update gate. This process enables GRUs to dynamically adjust their memory, balancing between retaining valuable historical information and adapting to new data.

These design features make GRUs well-suited for complex hydrological forecasting tasks where rapid analysis and handling of long-term dependencies are crucial. By balancing computational demands with the need for accurate and dynamic modeling, GRUs have become a popular choice in hydrological forecasting. Their suitability for applications in flood prediction and streamflow analysis, where they excel in modeling historical trends and adapting to new data patterns, further solidifies their value within this field.

### 3.4. Sequence-to-Sequence Model

In addition to LSTM and GRU networks, this study incorporates a variant of the Seq2Seq model, referenced from Xiang and Demir (2022), as a crucial baseline method. The Seq2Seq model, distinct for its encoder-decoder architecture, is designed to handle complex time-series forecasting tasks. It employs multiple TimeDistributed layers, culminating in a final dense layer to process and output predictions.

The Seq2Seq model operates with two distinct components: an encoder and a decoder. The encoder's role is to process the input time series data, such as historical rainfall, streamflow, and evapotranspiration from the past 72 hours. It converts this input into a context vector, which encapsulates essential temporal patterns and features, thus effectively summarizing the input data. GRUs are employed in this model for both the encoder and decoder, chosen for their efficacy in sequential data modeling and handling of long-range dependencies.

The decoder, on the other hand, is tasked with predicting future streamflow. It uses the context vector provided by the encoder, combined with additional future data inputs, to generate streamflow predictions for the next 120 hours. This process involves iteratively processing the context vector and the forecasted data, enabling the model to extend its predictions with each subsequent time step. To capture complex temporal dynamics, the Seq2Seq model employs multiple TimeDistributed layers, applying a consistent operation across each timestep of the output sequence.

The model's implementation concludes with a final dense layer, formatting the output sequence for the 120-hour streamflow predictions. For a detailed understanding of the Seq2Seq model's architecture and its application in hydrological forecasting, please refer to the study by Xiang and Demir (2022).

### 3.5. Transformer Model

The Transformer model, a groundbreaking development in neural network architecture, was introduced by Vaswani et al. (2017). Initially designed for machine translation, its innovative design has found widespread application in various domains, including time series forecasting. The key to the Transformer's effectiveness is its self-attention mechanism, which allows for a more efficient and comprehensive analysis of long input sequences compared to traditional recurrent layers.

Self-attention (Equation 4), the core feature of the Transformer, reinvents how input sequences are processed. Each element in the sequence is first transformed into three distinct vectors: queries (Q), keys (K), and values (V), each of dimension $d_{model}$. The self-attention scores are derived by performing dot-product operations between the query and key matrices. These scores are then scaled and passed through a softmax function to establish the relative importance of each element in relation to others. This process results in a new, weighted representation of the input sequence, allowing the Transformer to dynamically adjust each element's representation, considering the influence of all other elements in the sequence. This unique ability enables the model to capture long-range dependencies, which is crucial for accurate time series forecasting.

$$Attention(Q, K, V) = softmax\left(\frac{QK^T}{\sqrt{d_{model}}}\right)V \qquad \text{Eq. 4}$$

To enhance its pattern recognition capabilities, the Transformer employs multi-head attention. In this approach, the query, key, and value vectors are split into multiple segments. Each segment is processed by a separate attention head, allowing the model to concurrently focus on different aspects of the input sequence. The outputs from these attention heads are then concatenated and linearly transformed. Multi-head attention thus enriches the model's potential for capturing complex relationships within the sequence.

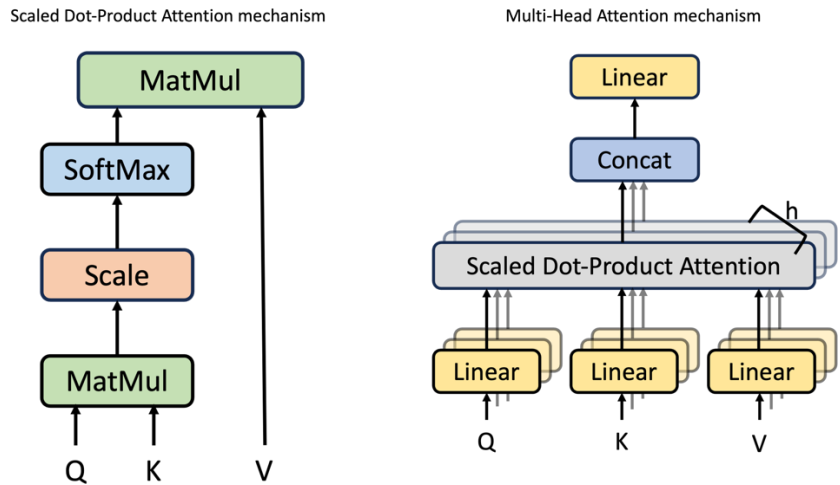

Figure 2: Visual Representation of Attention Mechanisms within the Transformer Architecture

A notable challenge in implementing self-attention is its lack of intrinsic positional awareness. Unlike recurrent neural networks, the Transformer does not process elements in sequence, which can lead to ambiguity regarding the order of elements. To overcome this, the Transformer incorporates static positional encoding. This encoding adds unique positional information to the input embeddings, ensuring that the model can distinguish between the positions of different elements within the sequence. Positional encoding is typically achieved using a specific formula involving sine and cosine functions of the positional index.

In our application of the Transformer model for hydrological forecasting, we have tailored the original framework to suit our specific needs. We introduced a linear embedding layer to adjust the size of the input features, preparing them for efficient processing by the Transformer's architecture. Additionally, we opted for a random variable for positional encoding, diverging from the traditional sine-cosine approach. This modification provides the model with necessary positional context while maintaining computational efficiency.

Our model also deviates in its structure, focusing on a single encoder layer equipped with eight attention heads. This configuration allows the model to simultaneously attend to different facets of the input sequence, enhancing its ability to process complex hydrological data. The model's size is set to 64, which strikes a balance between the model's complexity and computational demands. Furthermore, the encoder's internal feedforward network, with a dimension of 256, is designed to provide adequate capacity for processing and transforming features within the model. A GELU activation function is employed between two linear functions in the feedforward component, facilitating efficient non-linear transformations. The model is depicted in Figure 3.

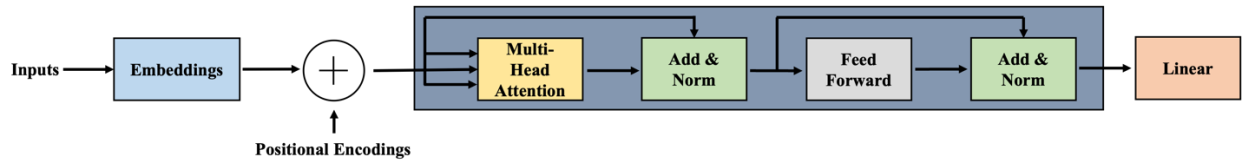

Figure 3: Visual Representation of the Transformer Architecture

In this study, the persistence, GRU, LSTM, and Transformer models were developed using PyTorch, while the Seq2Seq model was constructed with Keras. For comprehensive details on the implementation and architecture of the GRU, LSTM, and Seq2Seq models, please refer to the work by Demir et al. (2022). For the transformer model, throughout the training phase, Mean Absolute Error (MAE) was employed as the loss function, with Adam serving as the optimizer. We set the batch size to 512 and established a learning rate of 0.0001. To enhance model performance, the learning rate was halved if no improvement was observed over 10 epochs. Additionally, the training process was halted if no progress was noted for 20 consecutive epochs.

## 4. Results and Discussions

In this section, we present the findings of our investigation into 120-hour streamflow forecasting, with a particular focus on the performance of the Transformer model. This model is evaluated against four other models: three deep learning models, namely LSTM, GRU, and Seq2Seq, and the traditional Persistence method. Streamflow prediction is vital in fields like water resource management, environmental monitoring, and decision-making. Deep learning models have shown exceptional prowess in time-series forecasting, positioning them as suitable candidates for streamflow prediction. The Transformer model, with its innovative self-attention mechanism, is a relatively recent entrant in this domain and merits thorough exploration for its potential in capturing temporal dependencies in streamflow data.

Our comparative analysis utilizes four key metrics: Nash-Sutcliffe Efficiency (NSE), Kling–Gupta Efficiency (KGE), Pearson's r, and Normalized Root Mean Square Error (NRMSE). These metrics offer a comprehensive assessment of each model's predictive accuracy and the overall effectiveness of the Transformer model. We will provide a detailed analysis of these evaluation metrics and their significance in the context of streamflow forecasting. The subsequent sections will present a meticulous examination of the results from each model, shedding light on their individual strengths and limitations. Through this detailed evaluation, our goal is to discern the effectiveness of the Transformer model in 120-hour streamflow prediction and explore its implications for future research and practical applications.

## 4.1. Performance Metrics

In evaluating the efficacy of streamflow prediction models, it is essential to utilize robust performance metrics that accurately reflect the accuracy and reliability of the forecasts. For this purpose, this study employs four key metrics that are widely recognized in the field of hydrological modeling and streamflow forecasting: Nash-Sutcliffe Efficiency (NSE), Kling–Gupta Efficiency (KGE), Pearson's r and Normalized Root Mean Square Error (NRMSE). These metrics are chosen for their proven interpretability and comprehensive ability to assess various facets of model performance, as supported by previous research (Kratzert et al., 2018; Xiang and Demir, 2021; Liu et al., 2022).

First and foremost, Nash-Sutcliffe Efficiency (Equation 5) is a critical metric for gauging the predictive accuracy of hydrological models. As established in many studies (Krause et al., 2005; Arnold et al., 2012), NSE offers a quantifiable measure of the model's prediction capabilities in comparison to the observed streamflow data. Specifically, it evaluates the model's predictions relative to the average of the observed data. NSE values can range from negative infinity to a perfect score of 1, where 1 signifies an exact match between model predictions and observations. Scores above 0 indicate that the model predictions are superior to simply using the mean of the observed data. Conversely, negative NSE values suggest that using the mean of the observed data is more accurate than the model's predictions, denoting poor model performance. In the context of hydrological modeling, NSE values above 0.5 are generally considered acceptable, indicating a reasonable level of predictive accuracy (Arnold et al., 2012).

$$NSE = 1 - \frac{\sum(Y_i - \hat{Y}_i)^2}{\sum(Y_i - \bar{Y}_i)^2} \qquad \text{Eq. 5}$$

$Y_i = Observed\ streamflow\ value\ at\ time\ i$
$\hat{Y}_i = Predicted\ streamflow\ value\ at\ time\ i$
$\bar{Y}_i = Mean\ of\ all\ observations\ at\ time\ i$

Another key metric used in this study is the Kling-Gupta Efficiency (KGE) (Equation 6), which has become increasingly prominent in hydrological modeling. KGE offers a comprehensive evaluation of the model's performance by combining aspects of correlation, bias, and variability in a single metric. The formula for KGE is expressed as follows (Equation 6):

$$KGE = 1 - \sqrt{(r-1)^2 + (\alpha-1)^2 + (\beta-1)^2} \qquad \text{Eq. 6}$$

In the equation, $r$ is the correlation coefficient between observed and predicted streamflow, $\alpha$ is the ratio of the standard deviation of predicted streamflow to the standard deviation of observed streamflow, and $\beta$ is the ratio of the mean of predicted streamflow to the mean of observed streamflow. A KGE value of 1 indicates perfect model performance, reflecting an ideal balance between correlation, bias, and variability.

KGE provides a more holistic view of model performance compared to other metrics. It not only considers how well the predicted values match the observed data (correlation) but also evaluates the model's ability to accurately estimate the average and variability of the streamflow (bias and variability components). This makes KGE a valuable metric for assessing the overall suitability of a model in hydrological forecasting. Positive values of KGE are generally indicative of acceptable model performance, with higher values signifying better alignment between the model predictions and observed data.

Additionally, Pearson's correlation coefficient (Equation 7), commonly referred to as Pearson's r, serves as a crucial metric in our analysis. This statistical tool is employed to evaluate the linear relationship between the predicted streamflow values by the model and the actual observed streamflow data. Pearson's r effectively quantifies both the strength and the direction of this linear correlation. With its value ranging from -1 to 1, where 1 represents a perfect positive linear correlation, a higher positive value of Pearson's r indicates a model's enhanced reliability and accuracy. This is reflective of the model's capacity to closely follow the trends present in the observed data, thereby enabling precise predictions. By incorporating Pearson's r into our assessment, we can gauge the models' proficiency in capturing the observed streamflow variability throughout the 120-hour forecasting period.

$$r = \frac{\sum(Y_i - \bar{Y}_i)(\hat{Y}_i - \bar{\hat{Y}}_i)}{\sqrt{\sum(Y_i - \bar{Y}_i)^2}\sqrt{\sum(\hat{Y}_i - \bar{\hat{Y}}_i)^2}} \qquad \text{Eq. 7}$$

$Y_i = Observed\ streamflow\ value\ at\ time\ i$
$\hat{Y}_i = Predicted\ streamflow\ value\ at\ time\ i$
$\bar{Y}_i = Mean\ of\ all\ observations\ at\ time\ i$
$\bar{\hat{Y}}_i = Mean\ of\ all\ predicted\ values\ at\ time\ i$

The final metric used in this study to evaluate model performance is the Normalized Root Mean Square Error (NRMSE) (Equation 8). NRMSE quantifies the average magnitude of the errors between the predicted and observed streamflow values, normalized against the mean of the observed data. This metric offers a relative gauge of the model's accuracy, making it particularly useful for comparing performances across varying datasets. Given the diversity of locations in this study, NRMSE proves to be an apt choice for evaluation. The NRMSE value ranges between 0 and 1, where lower values signify superior model performance, denoting smaller errors in relation to the mean of the observed streamflow. Conversely, higher NRMSE values indicate larger errors and, consequently, lesser predictive accuracy.

These performance metrics play a crucial role in evaluating the effectiveness of our streamflow prediction models. They provide a quantitative basis for measuring how well these models can capture and replicate the complex patterns and dynamics inherent in streamflow data.

$$NRMSE = \frac{\sqrt{\frac{\sum_{i=1}^{n}(Y_i - \hat{Y}_i)^2}{\# \text{ of sample}}}}{\bar{Y}} \qquad \text{Eq. 8}$$

$Y_i = Observed\ streamflow\ value\ at\ time\ i$
$\hat{Y}_i = Predicted\ streamflow\ value\ at\ time\ i$
$\bar{Y} = Mean\ of\ all\ observations$

### 4.2. Experiment Results

In this part of the study, we detail the results from our extensive experiments focusing on predicting streamflow over a period of 120 hours. The central aim was to gauge the efficacy of the Transformer-based model in streamflow prediction, benchmarking its performance against three deep learning models (LSTM, GRU, and Seq2Seq) and the traditional Persistence approach. To measure the predictive accuracy of these models, we relied on four pivotal metrics widely recognized in hydrological modeling and streamflow forecasting: Nash-Sutcliffe Efficiency (NSE), Kling-Gupta Efficiency (KGE), Pearson's r, and Normalized Root Mean Square Error (NRMSE). These metrics are instrumental in evaluating the models' proficiency in accurately reflecting the intricacies of streamflow patterns.

In conducting our experiments, a key factor was the adaptation of input data dimensions to suit the specific requirements of the GRU, LSTM, and Transformer models. The input data comprised a blend of location-specific values, historical data, and forecast values. The historical data included 72 hours of precipitation, evapotranspiration, and discharge values, while the forecast data comprised predictions for 120 hours of precipitation and evapotranspiration. Consequently, the input data for these models was structured in two segments: one with the shape [batch size, 72, 10] for past values, and the other [batch size, 120, 9] for forecast values.

Table 3: Comparative performance summary of different models for 120-hour streamflow prediction on unified 125 stations (⬆ higher better, ⬇ lower better)

|  | NSE ⬆ | KGE ⬆ | R ⬆ | NRMSE ⬇ |
|---|---|---|---|---|
| **Persistence** | 0.864 | 0.922 | 0.931 | 0.824 |
| **SeqSeq** | 0.921 | 0.867 | 0.962 | 0.612 |
| **GRU** | 0.884 | 0.932 | 0.945 | 0.735 |
| **LSTM** | 0.864 | 0.903 | 0.939 | 0.793 |
| **Transformer** | 0.926 | 0.945 | 0.964 | 0.583 |

To effectively integrate these two sets of inputs for the models, it was necessary to introduce an additional dimension for the forecast values. For the LSTM and GRU models, we employed a zero-padding approach, extending the forecast values with zeros to match the dimensional requirements, in line with the benchmark paper's methodology. For the Transformer model, however, we adopted a persistence approach. This involved extending the historical values into the forecast period by repeating the last available data point, a technique shown effective in other work (Demiray et al., 2024). With this additional dimension in place, the past and forecast values were merged to form a unified input with dimensions [batch size, 192, 10] for use in the Transformer, GRU, and LSTM models.

In our comprehensive analysis, captured in Table 3 and Figures 4 and 5, we aggregated the 120-hour streamflow prediction results from all 125 stations, treating them as a unified dataset. This approach allowed us to observe the overall performance of each model as if it were predicting for a single, extensive location. For each hour, we calculated the NSE, KGE, and Pearson's r scores on the unified dataset and then determined their median values of 120 hours. The table summarizes these median values for NSE, KGE, and Pearson's r, as well as the NRMSE for each model.

According to Table 3, the Transformer model demonstrates proficiency in streamflow prediction, as evidenced by its high median scores in both NSE and KGE. This underlines the model's capability to accurately simulate and predict streamflow, maintaining a delicate balance between sensitivity to actual streamflow variations and error minimization. Additionally, its strong performance in Pearson's r and low NRMSE further corroborate its robustness and precision in predictions. In comparison, the Seq2Seq model, while showing a strong correlation with the observed data (as indicated by a high Pearson's r score), falls slightly short in NSE and KGE scores when compared to the Transformer. This suggests that while Seq2Seq is effective in capturing the linear relationship in streamflow data, it may not be as adept as the Transformer in balancing various aspects of model performance.

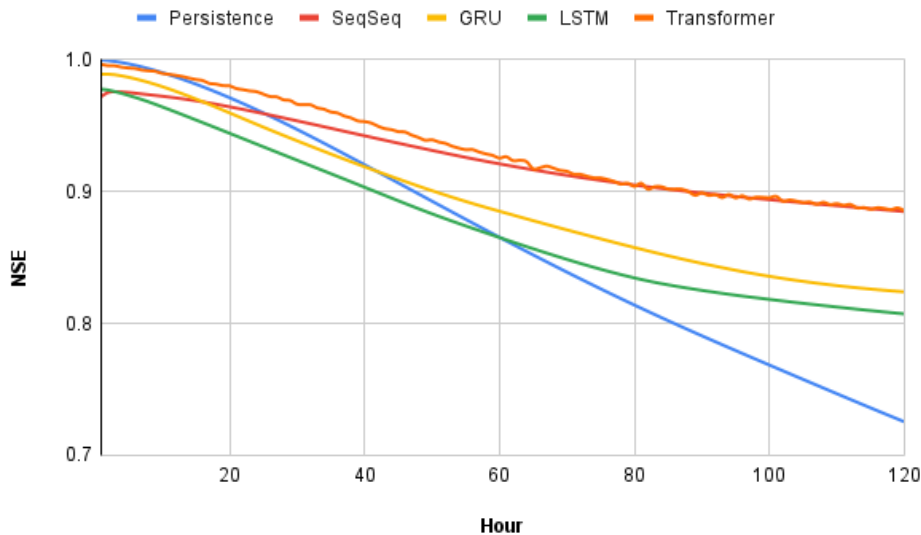

Figure 4: Hourly NSE scores 120-hour streamflow prediction on unified 125 stations

The GRU model's performance is noteworthy, particularly in terms of KGE, where it surpasses other models except the Transformer. This suggests its strength in balancing different components of model accuracy, such as correlation, bias, and variability. The LSTM model exhibits comparable NSE scores to the Persistence model, which serves as a baseline. This indicates that while LSTM is effective, it might not offer a significant advantage over simpler models in certain situations. Finally, the Persistence model's performance, particularly its high KGE score, is impressive considering its simplicity. This underscores the model's utility in certain forecasting scenarios where complex models might not provide significant additional benefits.

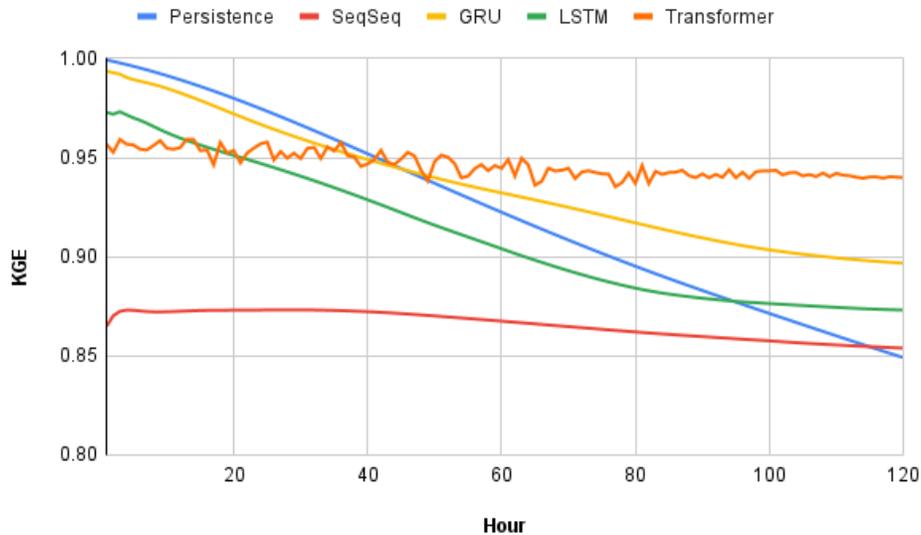

Figure 5: Hourly KGE scores 120-hour streamflow prediction on unified 125 stations

Figures 4 and 5 graphically represent the temporal changes in NSE and KGE scores, respectively, offering a visual representation of the models' ability to maintain accuracy over time. The Transformer model consistently maintains higher NSE and KGE scores across the 120-hour window, reinforcing its ability to sustain predictive performance over longer horizons. In contrast, the Persistence model, while starting strong, shows a more pronounced decline, particularly in NSE, as the forecast horizon extends. The Seq2Seq model, despite its robust Pearson's r value, exhibits variability in performance, particularly in KGE, suggesting limitations in balancing bias and variance when forecasting streamflow. These figures are instrumental in understanding the temporal stability of each model, with the Transformer model's lines depicting a slower decline in both NSE and KGE values, suggesting its robustness in handling the temporal complexity of streamflow prediction.

In order to assess the model performance across individual locations, we conducted a detailed location-wise analysis for the 120-hour streamflow prediction. For each of the 125 locations, we computed hourly NSE and KGE values, resulting in a distribution of these metrics for every hour of prediction. To synthesize this voluminous data, we took the median of the 125 NSE and KGE values at each hour, furnishing us with a representative performance indicator for that specific hour across all locations. This median-based aggregation method enabled us to mitigate the influence of any anomalous results that might be specific to a single location or time step. Figures 6 and 7 offer a visual representation of these 120 hours of NSE and KGE scores for each model. The summarization of these values is also presented in Table 4, which provides a snapshot of the models' capabilities by detailing the maximum, minimum, median, and mean values for both NSE and KGE metrics. The resulting median values reflect the models' general ability to capture the streamflow dynamics across diverse hydrological conditions and geographical variances, providing a concise summary of model effectiveness in typical settings and ensuring our conclusions are robust and reliable.

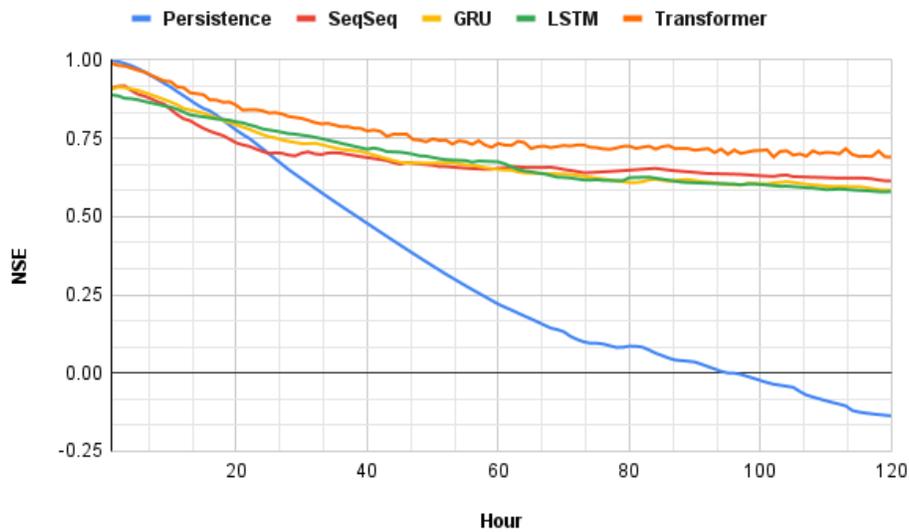

Figure 6: Hourly median NSE scores for 120-hour streamflow prediction across 125 stations.

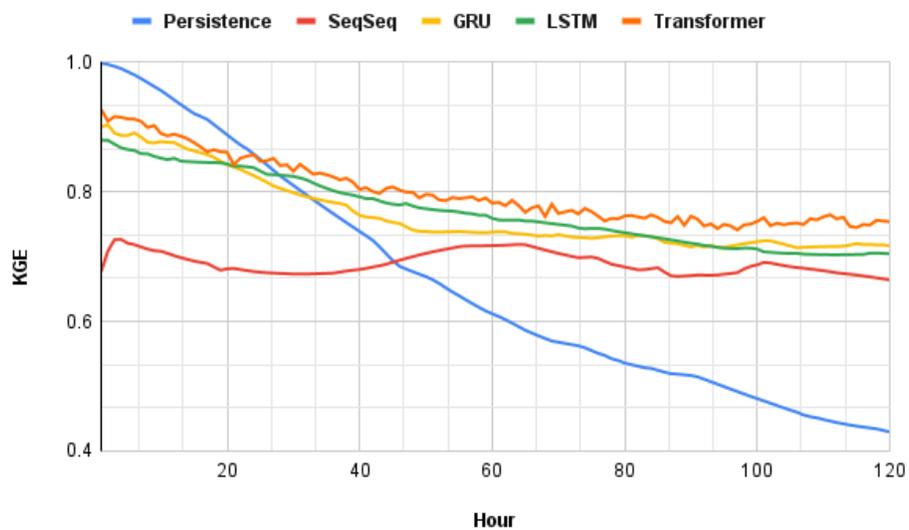

Figure 7: Hourly median KGE scores for 120-hour streamflow prediction across 125 stations.

Table 4: Statistical summary of hourly median NSE and KGE scores for 120-hour streamflow prediction across 125 stations (⬆ higher better)

|  | NSE ⬆ | | | | KGE ⬆ | | | |
| --- | --- | --- | --- | --- | --- | --- | --- | --- |
|  | **Min** | **Max** | **Median** | **Mean** | **Min** | **Max** | **Median** | **Mean** |
| **Persistence** | -0.136 | 0.998 | 0.215 | 0.321 | 0.429 | 0.998 | 0.609 | 0.659 |
| **SeqSeq** | 0.613 | 0.917 | 0.656 | 0.686 | 0.664 | 0.726 | 0.684 | 0.690 |
| **GRU** | 0.584 | 0.913 | 0.649 | 0.684 | 0.713 | 0.903 | 0.737 | 0.765 |
| **LSTM** | 0.578 | 0.888 | 0.670 | 0.686 | 0.702 | 0.879 | 0.758 | 0.770 |
| **Transformer** | 0.690 | 0.988 | 0.729 | 0.769 | 0.741 | 0.927 | 0.785 | 0.800 |

The results demonstrate a distinct superiority of the Transformer model in terms of both Nash-Sutcliffe Efficiency (NSE) and Kling-Gupta Efficiency (KGE), showcasing its robust predictive performance relative to the other models under consideration. Initially, the Persistence model registers the highest NSE and KGE values of 0.9987 and 0.9989, respectively, indicating a strong start. However, this model's performance sharply deteriorates, highlighting its limitations in extended forecasts. In stark contrast, the Transformer model's performance declines much more gradually, maintaining median NSE and KGE scores of 0.6902 and 0.7541 at the final prediction hour, which substantially exceeds the final scores of other models. This indicates not only the Transformer's sustained accuracy over time but also its generalizability across the 125 individual sensor locations.

When considering the aggregated results from the unified dataset, the Transformer model's proficiency is further affirmed, with the consistency of its performance across diverse conditions, as reflected in Table 3 and Figures 4 and 5. This consistent performance further validates the model's capability to generalize effectively across the spectrum of data variability and location-specific characteristics. The Seq2Seq, GRU, and LSTM models exhibit moderate performance in comparison, with the GRU model particularly notable for its KGE values, signifying its ability to balance accuracy components like correlation, bias, and variability. The aggregation of insights from both the individual sensor location predictions and the unified dataset results strengthens the case for the Transformer model as the preferred tool for streamflow prediction, adeptly navigating the complexities of hydrological patterns and offering a significant advancement over both traditional and contemporary modeling approaches.

In addition to conducting a location-wise analysis, we delved deeper into the performance metrics of each model. For each of the 125 sensor locations, we calculated NSE, KGE, and Pearson's r values hourly over a 120-hour forecast period. This provided a detailed performance profile for each model at every location, hour by hour. We then calculated the median of these 120 hourly values for each sensor, offering a stable and representative measure of performance by mitigating the impact of anomalies or extreme values. For the NRMSE, we assessed the performance over the entire 120-hour period as a whole for each sensor. The final aggregation step involved calculating a single median value for each metric across all 125 sensors, offering a comprehensive view of each model's overall performance across varied geographical and hydrological conditions. Table 5 presents these aggregated median values. For Tables 6 and 7, instead of aggregating median values of 125 sensors, we utilized the performance values of individual sensors to calculate the results shown in these tables.

Table 5: Aggregate median performance of models across 125 sensor locations (⬆ higher better, ⬇ lower better)

|  | NSE ⬆ | KGE ⬆ | R ⬆ | NRMSE ⬇ |
| --- | --- | --- | --- | --- |
| **Persistence** | 0.214741 | 0.609873 | 0.634093 | 1.065168 |
| **SeqSeq** | 0.667505 | 0.725735 | 0.871589 | 0.687117 |
| **GRU** | 0.660703 | 0.748974 | 0.857164 | 0.706444 |
| **LSTM** | 0.655519 | 0.768603 | 0.873331 | 0.709046 |
| **Transformer** | 0.746169 | 0.791779 | 0.884445 | 0.634105 |

Based on Table 5, the Transformer model showcases superior performance with the highest median values in all metrics, indicating its robustness and reliability in varied hydrological scenarios. Notably, its lower NRMSE value signifies a lower prediction error compared to other models. The Seq2Seq and GRU models also display commendable performance but are slightly outperformed by the Transformer. The Persistence model, simpler in its approach, shows lower median values in NSE and Pearson's r, reflecting its limitations in capturing complex streamflow dynamics.

Table 6: Number of stations where each model performed best across metrics (⬆ higher better, ⬇ lower better)

|  | NSE ⬆ | KGE ⬆ | R ⬆ | NRMSE ⬇ |
| --- | --- | --- | --- | --- |
| **Persistence** | 10 | 21 | 7 | 9 |
| **SeqSeq** | 21 | 9 | 29 | 17 |
| **GRU** | 19 | 20 | 22 | 19 |
| **LSTM** | 20 | 22 | 27 | 22 |
| **Transformer** | 55 | 53 | 40 | 58 |

Table 6 highlights the comparative effectiveness of each model at specific locations. The Transformer model consistently outperforms others in the majority of stations across all metrics, reinforcing its suitability for diverse forecasting scenarios. It leads significantly in NSE and KGE, demonstrating its effectiveness in capturing both accuracy and dynamics of streamflow predictions. Other models, including Seq2Seq, GRU, and LSTM, show a balanced distribution of best performances across different stations, suggesting their potential applicability in specific contexts.

Table 7: Count of sensor stations with NSE scores above 0.5 by model

|  | # NSE > 0.5 |
|---|---|
| **Persistence** | 45 |
| **SeqSeq** | 78 |
| **GRU** | 81 |
| **LSTM** | 85 |
| **Transformer** | 99 |

Table 7 focuses on the count of sensor stations where each model achieved an NSE score greater than 0.5. The Transformer model leads with 99 stations exceeding this threshold, validating its consistent and reliable predictive quality across a broad range of locations. The LSTM model follows closely, indicating its effective performance in many scenarios. The GRU and Seq2Seq models also demonstrate a significant number of stations with NSE scores above 0.5, while the Persistence model, with its simpler approach, understandably lags in this aspect.

As we draw the results section of this study to a close, it becomes clear that the Transformer model distinctly outperforms its counterparts—LSTM, GRU, Seq2Seq, and the traditional Persistence approach—in predicting streamflow over a 120-hour period. The comprehensive analyses, encapsulated in Tables 3, 4, 5, 6, and 7, and Figures 4, 5, 6, and 7, have consistently highlighted the Transformer model's superior performance. Notably, its ability to maintain higher median scores in Nash-Sutcliffe Efficiency (NSE), Kling-Gupta Efficiency (KGE), and Pearson's r, coupled with the lowest Normalized Root Mean Square Error (NRMSE), underscores its robustness and precision in diverse hydrological scenarios.

The model's effectiveness is further affirmed by its dominant performance across a majority of the 125 sensor locations, signifying its adaptability and reliability in various forecasting contexts. These findings not only validate the Transformer model's capability in balancing accuracy, bias, and variability in streamflow predictions but also mark it as a significant advancement over both traditional and contemporary modeling approaches.

## 5. Conclusion

In this study, we have explored the effectiveness of a generalized Transformer model in the context of 120-hour streamflow prediction, utilizing a comprehensive dataset encompassing 125 sensor locations across Iowa, USA. This approach represents a departure from traditional methods, as we developed a generalized model trained on data from all these locations simultaneously, rather than creating separate models for each site. By integrating 72 hours of historical data on precipitation, evapotranspiration, and discharge, along with location-specific attributes, the model has been tailored to capture the complex dynamics of hydrological systems effectively.

Our analysis, grounded in four key metrics - Nash-Sutcliffe Efficiency (NSE), Kling-Gupta Efficiency (KGE), Pearson's r, and Normalized Root Mean Square Error (NRMSE) - has revealed the Transformer model's superior capability in streamflow prediction compared to LSTM, GRU, Seq2Seq, and the Persistence approach. The model demonstrated higher median scores in NSE and KGE, alongside the lowest NRMSE values, underscoring its precision and reliability in diverse hydrological scenarios. The Transformer model's predominant performance across a majority of the sensor locations further accentuates its adaptability and applicability in various forecasting contexts. This study not only underscores the model's efficacy in streamflow prediction but also contributes significantly to the field of hydrological modeling, showcasing the potential of advanced deep learning techniques in enhancing prediction accuracy and efficiency.

Looking ahead, there are numerous opportunities for extending this research. Future studies could explore the application of the Transformer model in different hydrological settings, integrate a broader range of environmental variables, and assess its scalability for larger geographical areas. Further development and refinement of the model's architecture and training process could also enhance its predictive capabilities. Such advancements will be crucial for improving water resource management strategies and strengthening our preparedness for hydrological extremes in the face of evolving climate conditions.

## 6. References

Ahmed, A.M., Deo, R.C., Feng, Q., Ghahramani, A., Raj, N., Yin, Z. and Yang, L. (2021). Deep learning hybrid model with Boruta-Random forest optimiser algorithm for streamflow forecasting with climate mode indices, rainfall, and periodicity. Journal of Hydrology, 599, p.126350.

Alabbad, Y., & Demir, I. (2022). Comprehensive flood vulnerability analysis in urban communities: Iowa case study. International journal of disaster risk reduction, 74, 102955.

Arnold, J., 1994. SWAT-soil and water assessment tool.

Banholzer, S., Kossin, J., & Donner, S. (2014). The impact of climate change on natural disasters. In Reducing disaster: Early warning systems for climate change (pp. 21-49). Springer, Dordrecht.

Bayar, S., Demir, I., & Engin, G. O. (2009). Modeling leaching behavior of solidified wastes using back-propagation neural networks. Ecotoxicology and environmental safety, 72(3), 843-850.

Beven, K.J. and Kirkby, M.J., 1979. A physically based, variable contributing area model of basin hydrology. Hydrological sciences journal, 24(1), pp.43-69.

Castangia, M., Grajales, L.M.M., Aliberti, A., Rossi, C., Macii, A., Macii, E. and Patti, E., 2023. Transformer neural networks for interpretable flood forecasting. Environmental Modelling & Software, 160, p.105581.

Chen, Z., Lin, H. and Shen, G., 2023. TreeLSTM: A spatiotemporal machine learning model for rainfall-runoff estimation. Journal of Hydrology: Regional Studies, 48, p.101474.

Davenport, F. V., Burke, M., & Diffenbaugh, N. S. (2021). Contribution of historical precipitation change to US flood damages. Proceedings of the National Academy of Sciences, 118(4).

Devia, G.K., Ganasri, B.P. and Dwarakish, G.S., 2015. A review on hydrological models. Aquatic procedia, 4, pp.1001-1007.

Demir, I., & Beck, M. B., 2009, April. GWIS: a prototype information system for Georgia watersheds. In Georgia Water Resources Conference: Regional Water Management Opportunities, UGA, Athens, GA, US.

Demir, I., Xiang, Z., Demiray, B. and Sit, M., 2022. WaterBench-Iowa: a large-scale benchmark dataset for data-driven streamflow forecasting. Earth system science data, 14(12), pp.5605-5616.

Demiray, B.Z., Sit, M. and Demir, I., 2023. EfficientTempNet: Temporal Super-Resolution of Radar Rainfall. arXiv preprint arXiv:2303.05552.

Demiray, B.Z., Sit, M., Mermer, O. and Demir, I., 2024. Enhancing Hydrological Modeling with Transformers: A Case Study for 24-Hour Streamflow Prediction. Water Science & Technology 89 (9), 2326-2341.

Diffenbaugh, N. S., Singh, D., Mankin, J. S., Horton, D. E., Swain, D. L., Touma, D., ... & Rajaratnam, B. (2017). Quantifying the influence of global warming on unprecedented extreme climate events. Proceedings of the National Academy of Sciences, 114(19), 4881-4886.

Granata, F., Gargano, R. and De Marinis, G., 2016. Support vector regression for rainfall-runoff modeling in urban drainage: A comparison with the EPA's storm water management model. Water, 8(3), p.69.

Guo, Y., Yu, X., Xu, Y.P., Chen, H., Gu, H. and Xie, J., 2021. AI-based techniques for multi-step streamflow forecasts: application for multi-objective reservoir operation optimization and performance assessment. Hydrology and Earth System Sciences, 25(11), pp.5951-5979.

Hochreiter, S. and Schmidhuber, J., 1997. Long short-term memory. Neural computation, 9(8), pp.1735-1780.

Honorato, A.G.D.S.M., Silva, G.B.L.D. and Guimaraes Santos, C.A., 2018. Monthly streamflow forecasting using neuro-wavelet techniques and input analysis. Hydrological Sciences Journal, 63(15-16), pp.2060-2075.

Ibrahim, K.S.M.H., Huang, Y.F., Ahmed, A.N., Koo, C.H. and El-Shafie, A., 2022. A review of the hybrid artificial intelligence and optimization modelling of hydrological streamflow forecasting. Alexandria Engineering Journal, 61(1), pp.279-303.

IPCC, 2022: Climate Change 2022: Impacts, Adaptation, and Vulnerability. Contribution of Working Group II to the Sixth Assessment Report of the Intergovernmental Panel on Climate Change [H.-O. Pörtner, D.C. Roberts, M. Tignor, E.S. Poloczanska, K. Mintenbeck, A. Alegría, M. Craig, S. Langsdorf, S. Löschke, V. Möller, A. Okem, B. Rama (eds.)]. Cambridge University Press. Cambridge University Press, Cambridge, UK and New York, NY, USA, 3056 pp., doi:10.1017/9781009325844.

Krajewski, W.F., Ghimire, G.R., Demir, I. and Mantilla, R., 2021. Real-time streamflow forecasting: AI vs. Hydrologic insights. Journal of Hydrology X, 13, p.100110.

Kratzert, F., Klotz, D., Brenner, C., Schulz, K. and Herrnegger, M., 2018. Rainfall–runoff modelling using long short-term memory (LSTM) networks. Hydrology and Earth System Sciences, 22(11), pp.6005-6022.

Lee, T.H. and Georgakakos, K.P., 1996. Operational Rainfall Prediction on Meso-γ Scales for Hydrologic Applications. Water Resources Research, 32(4), pp.987-1003.

Li, Z., & Demir, I., 2022. A comprehensive web-based system for flood inundation map generation and comparative analysis based on height above nearest drainage. Science of The Total Environment, 828, 154420.

Li, Z., & Demir, I., 2024. Better localized predictions with Out-of-Scope information and Explainable AI: One-Shot SAR backscatter nowcast framework with data from neighboring region. ISPRS Journal of Photogrammetry and Remote Sensing, 207, 92-103.

Lin, T., Wang, Y., Liu, X. and Qiu, X., 2022. A survey of transformers. AI Open.

Liu, C., Liu, D. and Mu, L., 2022. Improved transformer model for enhanced monthly streamflow predictions of the Yangtze River. IEEE Access, 10, pp.58240-58253.

Mosavi, A., Ozturk, P. and Chau, K.W., 2018. Flood prediction using machine learning models: Literature review. Water, 10(11), p.1536.

Munich Re. (2022). Hurricanes, cold waves, tornadoes: Weather disasters in USA dominate natural disaster losses in 2021. https://www.munichre.com/en/company/media-relations/media -information-and-corporate-news/media-information/2022/natural-disaster-losses-2021.html.

NDRCC. (2021). 2020 Global Natural Disaster Assessment Report. https://reliefweb.int/report/china/2020-global-natural-disaster-assessment-report

Nguyen, T.T.H., Vu, D.Q., Mai, S.T. and Dang, T.D., 2023. Streamflow Prediction in the Mekong River Basin Using Deep Neural Networks. IEEE Access, 11, pp.97930-97943.

NOAA National Centers for Environmental Information (NCEI). (2022). US billion-dollar weather and climate disasters. https://www.ncei.noaa.gov/access/monitoring/billions/, DOI:10.25921/stkw-7w73

Ren-Jun, Z., 1992. The Xinanjiang model applied in China. Journal of hydrology, 135(1-4), pp.371-381.

Salas, J.D., Markus, M. and Tokar, A.S. (2000). Streamflow forecasting based on artificial neural networks. Artificial neural networks in hydrology, pp.23-51.

Sharma, P. and Machiwal, D. eds., 2021. Advances in streamflow forecasting: from traditional to modern approaches. Elsevier.

Sit, M., Demiray, B. and Demir, I., 2021. Short-term hourly streamflow prediction with graph convolutional GRU networks. arXiv preprint arXiv:2107.07039.

Sit, M., Demiray, B.Z. and Demir, I., 2022. A systematic review of deep learning applications in streamflow data augmentation and forecasting. EarthArxiv, 3617. https://doi.org/10.31223/X5HM08

Sit, M., Demiray, B.Z. and Demir, I., 2023. Spatial Downscaling of Streamflow Data with Attention Based Spatio-Temporal Graph Convolutional Networks. EarthArxiv, 5227. https://doi.org/10.31223/X5666M

Sit, M., Seo, B. C., Demiray, B., & Demir, I., 2024. EfficientRainNet: Leveraging EfficientNetV2 for memory-efficient rainfall nowcasting. Environmental Modelling & Software, 176, 106001.

Strauss, B. H., Kopp, R. E., Sweet, W. V., and Bittermann, K. (2016). Unnatural coastal floods: Sea level rise and the human fingerprint on US floods since 1950. Climate Central.

Tabari, H. (2020). Climate change impact on flood and extreme precipitation increases with water availability. Scientific reports, 10(1), 1-10.

Trenberth, K. E., Cheng, L., Jacobs, P., Zhang, Y., & Fasullo, J. (2018). Hurricane Harvey links to ocean heat content and climate change adaptation. Earth's Future, 6(5), 730-744.

UNESCO. (2023). The United Nations world water development report 2023: partnerships and cooperation for water. UN.

Vaswani, A., Shazeer, N., Parmar, N., Uszkoreit, J., Jones, L., Gomez, A.N., Kaiser, Ł. and Polosukhin, I., 2017. Attention is all you need. Advances in neural information processing systems, 30.

World Meteorological Organization (WMO). (2021). The Atlas of Mortality and Economic Losses from Weather, Climate and Water Extremes (1970–2019).

Wu, H., Xu, J., Wang, J. and Long, M., 2021. Autoformer: Decomposition transformers with auto-correlation for long-term series forecasting. Advances in Neural Information Processing Systems, 34, pp.22419-22430.

Xiang, Z., Demir, I., Mantilla, R., & Krajewski, W. F., 2021. A regional semi-distributed streamflow model using deep learning. EarthArxiv, 2152. https://doi.org/10.31223/X5GW3V

Xiang, Z. and Demir, I., 2022. Fully distributed rainfall-runoff modeling using spatial-temporal graph neural network. EarthArxiv, 3018. https://doi.org/10.31223/X57P74

Xu, Y., Lin, K., Hu, C., Wang, S., Wu, Q., Zhang, L. and Ran, G., 2023. Deep transfer learning based on transformer for flood forecasting in data-sparse basins. Journal of Hydrology, 625, p.129956.

Yan, J., Jin, J., Chen, F., Yu, G., Yin, H. and Wang, W., 2018. Urban flash flood forecast using support vector machine and numerical simulation. Journal of Hydroinformatics, 20(1), pp.221-231.

Yaseen, Z.M., El-Shafie, A., Jaafar, O., Afan, H.A. and Sayl, K.N. (2015). Artificial intelligence based models for streamflow forecasting: 2000–2015. Journal of Hydrology, 530, pp.829-844.

Yaseen, Z.M., Ebtehaj, I., Bonakdari, H., Deo, R.C., Mehr, A.D., Mohtar, W.H.M.W., Diop, L., El-Shafie, A. and Singh, V.P. (2017). Novel approach for streamflow forecasting using a hybrid ANFIS-FFA model. Journal of Hydrology, 554, pp.263-276.

Yaseen, Z.M., Awadh, S.M., Sharafati, A. and Shahid, S. (2018). Complementary data-intelligence model for river flow simulation. Journal of Hydrology, 567, pp.180-190.

Yaseen, Z.M., Sulaiman, S.O., Deo, R.C. and Chau, K.W., 2019. An enhanced extreme learning machine model for river flow forecasting: State-of-the-art, practical applications in water resource engineering area and future research direction. Journal of Hydrology, 569, pp.387-408.

Yildirim, E., Just, C., & Demir, I., 2022. Flood risk assessment and quantification at the community and property level in the State of Iowa. International journal of disaster risk reduction, 77, 103106.

Zhou, H., Zhang, S., Peng, J., Zhang, S., Li, J., Xiong, H. and Zhang, W., 2021, May. Informer: Beyond efficient transformer for long sequence time-series forecasting. In Proceedings of the AAAI conference on artificial intelligence (Vol. 35, No. 12, pp. 11106-11115).

Zhou, T., Ma, Z., Wen, Q., Wang, X., Sun, L. and Jin, R., 2022, June. Fedformer: Frequency enhanced decomposed transformer for long-term series forecasting. In International Conference on Machine Learning (pp. 27268-27286). PMLR.